\documentclass{article}
\usepackage{spconf,amsmath,epsfig}
\usepackage[table]{xcolor}
\let\OLDthebibliography\thebibliography
\renewcommand\thebibliography[1]{
  \OLDthebibliography{#1}
  \setlength{\parskip}{0pt}
  \setlength{\itemsep}{0pt plus 0.3ex}
}

\begin{document}\sloppy


\title{GDGS: 3D Gaussian Splatting Via Geometry-Guided Initialization And Dynamic  Density Control}
%
\name{
Xingjun Wang, Lianlei Shan
}
\address{Tsinghua University}

\maketitle

\begin{abstract}
  We propose a method to enhance 3D Gaussian Splatting (3DGS)~\cite{Kerbl2023}, addressing challenges in initialization, optimization, and density control. Gaussian Splatting is an alternative for rendering realistic images while supporting real-time performance, and it has gained popularity due to its explicit 3D Gaussian representation. However, 3DGS heavily depends on accurate initialization and faces difficulties in optimizing unstructured Gaussian distributions into ordered surfaces, with limited adaptive density control mechanism proposed so far. Our first key contribution is a geometry-guided initialization to predict Gaussian parameters, ensuring precise placement and faster convergence. We then introduce a surface-aligned optimization strategy to refine Gaussian placement, improving geometric accuracy and aligning with the surface normals of the scene. Finally, we present a dynamic adaptive density control mechanism that adjusts Gaussian density based on regional complexity, for visual fidelity. These innovations enable our method to achieve high-fidelity real-time rendering and significant improvements in visual quality, even in complex scenes. Our method demonstrates comparable or superior results to state-of-the-art methods, rendering high-fidelity images in real time.
\end{abstract}
\begin{keywords}
  3D Gaussian Splatting (3DGS), Novel view synthesis, Real-time rendering, Structure-from-Motion (SfM)
\end{keywords}
\section{Introduction}
\label{sec:intro}

Novel view synthesis is a fundamental task in computer vision and graphics. 3D Gaussian Splatting (3DGS) ~\cite{Kerbl2023}has emerged as a cutting-edge approach for capturing and rendering 3D scenes from novel perspectives. Unlike NeRFs~\cite{Mildenhall2020}, which rely on MLPs which are computationally intensive and resource-demanding, 3DGS directly models scenes using 3D Gaussians. This method optimizes Gaussian positions, orientations, appearances, and alpha blending to represent the scene’s geometry and appearance efficiently. 

Current 3DGS methods encode scene geometry and appearance by optimizing parameters such as position, covariance, and color of 3D Gaussians. Despite their flexibility, these methods face challenges in aligning unstructured Gaussian distributions into ordered surfaces. Additionally, uniform treatment of all image regions leads to inefficiencies, as high-detail or close-up areas demand finer sampling, while simpler areas incur unnecessary computational costs. 

We introduce three key innovations. First, an improved geometric initialization strategy generates a structured and reliable point cloud, outperforming point cloud from Structure-from-Motion. Second, surface normals are aligned with planes to further improve geometric accuracy. Third, a novel adaptive density control (ADC) mechanism leverages dynamic resolution to determine regions requiring additional Gaussians. Unlike current approaches that delete overly transparent or camera-proximal points and clone large high-gradient Gaussians, this method uses fixed region segmentation to assess detail needs based on Gaussian density and gradient magnitude. In regions requiring adjustment, evenly distributed Gaussians are increased via cloning, while uneven distributions are refined by modifying regional loss functions to improve Gaussian allocation.

Our proposed approach results in images with minimal pixel-level distortion, successfully preserving overall structural integrity. The method ensures a higher degree of structural similarity while reducing pixel-wise discrepancies, achieving superior accuracy in capturing intricate details of the scene, especially in high-detail regions. This leads to improved rendering quality and enhanced performance for real-time rendering tasks.

\section{Related Works}

Our research builds on 3D Gaussian Splatting (3DGS)~\cite{Kerbl2023}. We discuss related works in traditional scene reconstruction, neural rendering, and point-based rendering.

Early scene reconstruction methods leveraged light fields for novel-view synthesis ~\cite{Gortler1996, Levoy1996}, progressing to unstructured captures ~\cite{Buehler2001}. Structure-from-Motion (SfM) ~\cite{Snavely2006}introduced sparse point clouds for visualizing 3D space, further enhanced by Multi-View Stereo (MVS) ~\cite{Goesele2007} for dense reconstructions. These methods achieved compelling results in tasks such as re-projection-based view synthesis ~\cite{Chaurasia2013, Eisemann2008,Hedman2018,Kopanas2021}. However, challenges remained with artifacts like unreconstructed regions and over-reconstructed geometry.

Neural rendering algorithms ~\cite{Tewari2022} have significantly reduced these issues, offering superior performance without the overhead of storing all input images on the GPU. These methods have established neural rendering as a robust alternative to traditional approaches for diverse applications.

Point-based rendering ~\cite{Grossman1998, Botsch2005, Pfister2000, Zwicker2001b} provides an efficient way to handle unstructured geometry but often suffers from discontinuities and aliasing. Differentiable point-based techniques ~\cite{Aliev2020,Ruckert2022} have incorporated neural features for improved performance, but their reliance on MVS-derived geometry limits their robustness in complex scenes. Pulsar \cite{Lassner2021}introduced fast sphere rasterization, inspiring the efficient rasterization techniques used in 3DGS.

NeRF ~\cite{Mildenhall2020} marked a significant advancement in novel-view synthesis by rendering 3D views through ray integration of 2D data. NeRF encodes positional information using positional encoding to improve spatial understanding and utilizes hierarchical volume sampling for enhanced rendering through multi-level sampling. NeRF trains a MLP to predict the density and radiance at any 3D point.

While subsequent models have extended NeRF’s capabilities for dynamic scenes, reduced data requirements, and accelerated training using external tools like hash grids ~\cite{Muller2022}, its reliance on computationally expensive ray-based querying limits its rendering speed, making it unsuitable for real-time applications.

3DGS ~\cite{Kerbl2023} introduces an explicit scene representation using 3D Gaussian primitives, which offers substantial advantages for real-time novel-view synthesis. Unlike NeRF’s implicit volumetric representation, 3DGS directly models scenes with explicit 3D Gaussians. These are rasterized into image space using a fast and differentiable CUDA-based algorithm, enabling real-time rendering at high resolutions. SuGaR ~\cite{Guedon2024} aligns Gaussian splats with surface normals, enhancing the fidelity of 3D mesh reconstruction and enabling high-quality rendering. Multi-Scale 3D Gaussian Splatting ~\cite{Yan2024} introduces a multi-scale approach to Gaussian splatting, ensuring anti-aliased rendering by dynamically adapting splat density and scale based on scene complexity. And others also do some meaningful works in segmentation \cite{densenet,uhrsnet,tgrs1,decouple,mbnet,tgrs2,liminglong,zhaoyuzhong,acmmm,boosting1,data,zhaoguiqin,lifelong,cognitive,dlnet,rs2,fusing,ldnet,organizing,edge,binary,energy,boosting,synthetic,dynrsl,flexdataset,gmm,llmcot,geogrambench,geolocsft,f2net}.

\section{Method}
\subsection{Overview}

\begin{figure}[htb]
\begin{minipage}[b]{1.0\linewidth}
  \centering
  \includegraphics[scale=0.30]{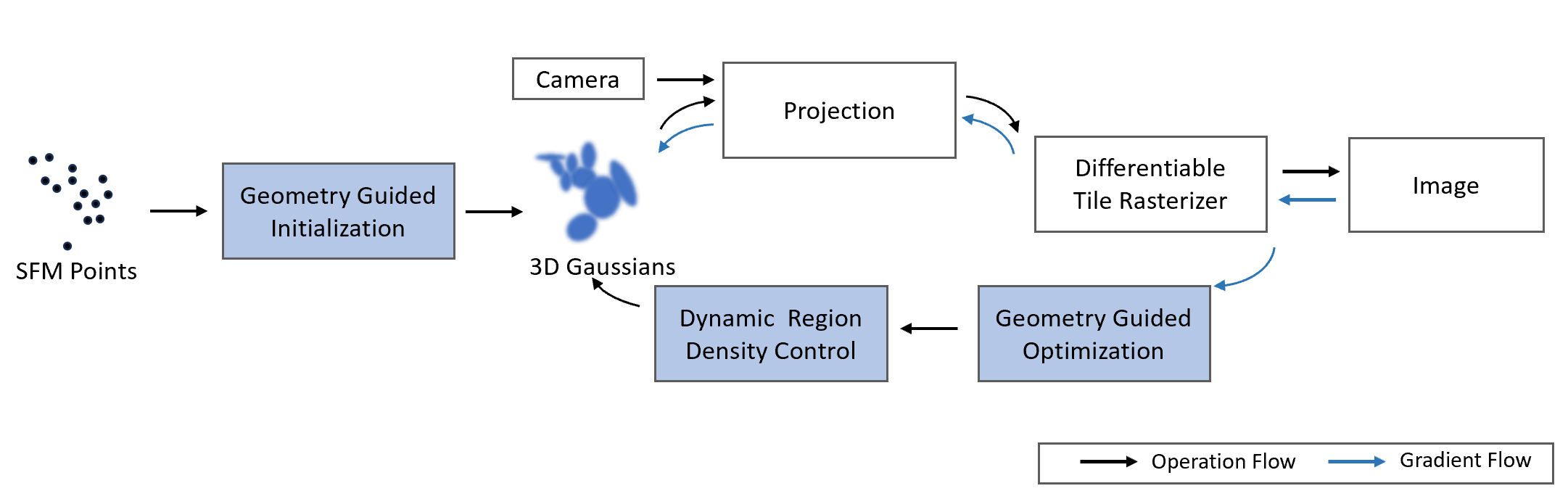}
  \caption{Optimization process begins with Structure-from-Motion (SfM) points, points are utilized in a geometry-guided initialization phase to accurately position 3D Gaussians according to the scene's geometry. The initialized Gaussians, undergo further optimization to ensure they align with the surface normals, enhancing geometric accuracy.
 Following this, dynamic region density control is applied, adjusting the Gaussians' density across the scene to improve rendering efficiency and quality.}
\label{fig:p1}
\end{minipage}
\end{figure}
Our research builds upon the 3D Gaussian Splatting (3DGS) framework~\cite{Kerbl2023}  by introducing additional steps to enhance its initialization, optimization, and adaptive density control as shown in Figure~\ref{fig:p1}. Starting with input images and camera parameters calibrated using Structure-from-Motion (SfM) ~\cite{Snavely2006}, we use the resulting sparse point cloud as input to a Multi-Layer Perceptron (MLP), which predicts initial Gaussian parameters—positions, covariance matrices, and opacity. This data-driven initialization replaces random parameter generation, ensuring a more precise starting point and accelerating convergence. The initialized Gaussians are then optimized to align with surface normals extracted from mesh data, ensuring they remain external to the object while capturing fine surface details. This optimization is guided by a composite loss function that balances distance, direction, and surface fitting losses. Additionally, our dynamic Adaptive Density Control (ADC) divides the scene into fixed 3D grid regions, dynamically refining Gaussian density based on regional complexity. High-complexity regions are enriched by cloning Gaussians, while simpler regions undergo loss-based dispersion to maintain uniformity and avoid redundancy. The optimized Gaussians are rendered through a tile-based rasterizer achieving real-time performance.

\subsection{Geometry Guided Adaptive Optimization}
\begin{figure}[htb]
\begin{minipage}[b]{1.0\linewidth}
  \centering
  \includegraphics[scale=0.47]{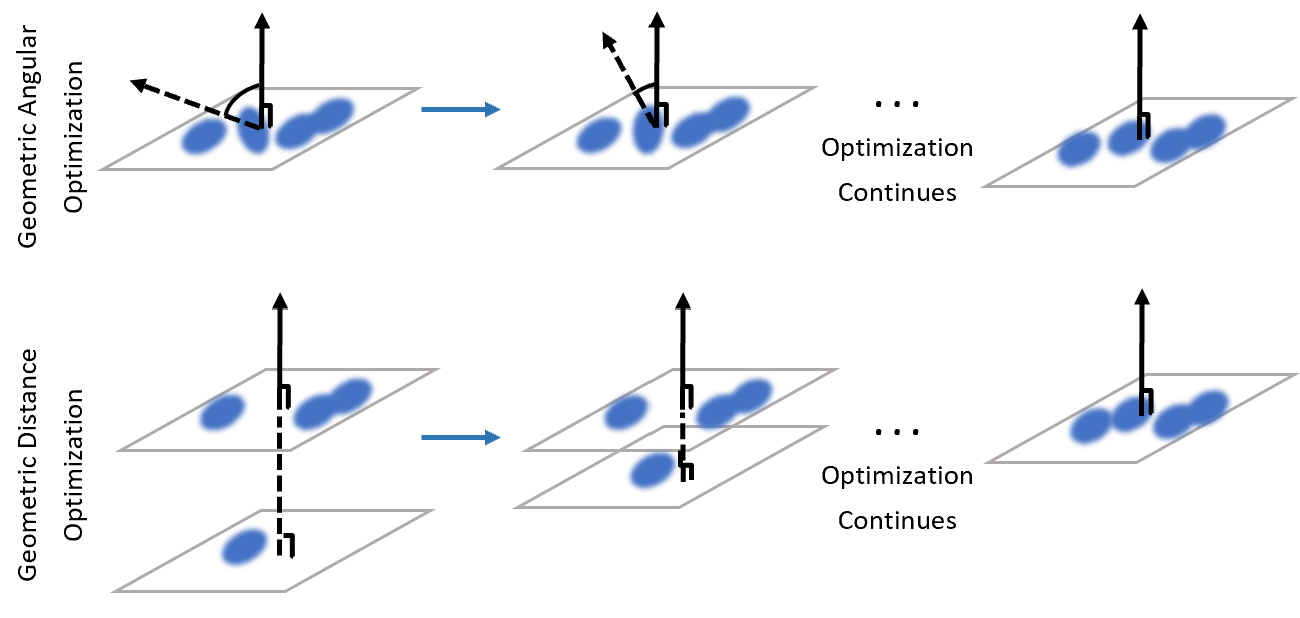}
  \caption{Following initialization, positioning logic is applied to adjust the point orientation and distance, by setting training objectives and defining a loss function specifically for the orientation and position convergence criteria.}
  \label{fig:p2}
\end{minipage}
\end{figure}

Our goal is to achieve high-quality initialization and optimization of 3D Gaussians for novel-view synthesis by leveraging structured data. This is accomplished in two distinct stages: an adaptive initialization stage, which ensures precise placement of Gaussians, and a dynamic optimization stage, which refines their positions and orientations for accurate reconstruction as shown in Figure~\ref{fig:p2}. Together, these stages address the challenges of reconstructing 3D scenes with both precision and efficiency.

The adaptive initialization process begins by utilizing structured outputs such as camera intrinsics, extrinsics, and a sparse point cloud. These outputs provide the foundation for predicting Gaussian centers (\(\mu\)) using a Multi-Layer Perceptron (MLP). Sparse point clouds capture the geometric structure of the scene but lack reliable surface normals. To overcome this limitation, the scene is modeled as a set of 3D Gaussians, each defined by a center position (\(\mu\)) and covariance matrix (\(\Sigma\)). The MLP, trained on camera calibration and point cloud data, predicts Gaussian centers that closely align with the scene geometry, significantly improving initialization accuracy and optimization convergence.

The MLP architecture takes 3D coordinates from the point cloud as input and processes them through multiple layers with nonlinear activations (e.g., ReLU) to model complex relationships. The output layer directly predicts the Gaussian center positions (\(\mu\)), which serve as the starting points for subsequent optimization. During training, a loss function evaluates the deviation of predicted Gaussian centers from their ground truth:
\begin{equation}
    \mathcal{L}_{\text{init}} = \frac{1}{N} \sum_{i=1}^N \| \mu_i^{\text{pred}} - \mu_i^{\text{gt}} \|^2,
\end{equation}
where \(\mu_i^{\text{pred}}\) and \(\mu_i^{\text{gt}}\) represent the predicted and ground truth centers, respectively. The training process employs normalized input features to ensure uniform contributions across dimensions, preventing gradient saturation. Data augmentation, such as Gaussian noise addition, enhances robustness and generalization.

Following initialization, dynamic optimization refines Gaussian placement to better align with surface geometry and capture scene complexity. Each Gaussian \(i\) is iteratively adjusted to position it slightly outside the object surface, perpendicular to its normal vector. The target position is:
\begin{equation}
    \mu_i^{\text{target}} = \mu_i^{\text{mesh}} + d \cdot \mathbf{N},
\end{equation}
where \(\mu_i^{\text{mesh}}\) is the nearest point on the surface mesh, \(d\) is a positive offset, and \(\mathbf{N}\) is the surface normal. The iterative update rule adjusts Gaussian centers toward their targets:
\begin{equation}
    \mu_i \leftarrow \mu_i - \eta \nabla_{\mu_i} \mathcal{L},
\end{equation}
where \(\eta\) is the learning rate, and \(\mathcal{L}\) represents the total loss.

The optimization process is governed by a composite loss function:
\begin{equation}
    \mathcal{L} = \lambda_d \mathcal{L}_{\text{dist}} + \lambda_a \mathcal{L}_{\text{align}} + \lambda_s \mathcal{L}_{\text{surface}},
\end{equation}
where \(\lambda_d, \lambda_a, \lambda_s\) are hyperparameters balancing the components. The distance loss evaluates deviations from the target:
\begin{equation}
    \mathcal{L}_{\text{dist}} = \frac{1}{N} \sum_{i=1}^N \|\mu_i - \mu_i^{\text{target}}\|^2.
\end{equation}
The alignment loss ensures Gaussian orientations align with surface normals:
\begin{equation}
    \mathcal{L}_{\text{align}} = \frac{1}{N} \sum_{i=1}^N (1 - \cos \theta_i),
\end{equation}
where \(\cos \theta_i = \frac{\mathbf{U}_i \cdot \mathbf{N}_i}{\|\mathbf{U}_i\| \|\mathbf{N}_i\|}\), and \(\mathbf{U}_i\) is the Gaussian’s orientation vector. The surface fitting loss measures alignment with the mesh:
\begin{equation}
    \mathcal{L}_{\text{surface}} = \frac{1}{N} \sum_{i=1}^N \sum_{j \in \mathcal{N}(i)} \|\mu_i - \mu_j^{\text{mesh}}\|^2,
\end{equation}
where \(\mathcal{N}(i)\) denotes neighboring mesh points of Gaussian \(i\).

Optimization continues until the total loss falls below a predefined threshold, ensuring both efficiency and accuracy. Dynamic learning rates adapt to the convergence rate, further stabilizing the process. By integrating adaptive initialization with dynamic optimization, this framework ensures precise Gaussian placement and enhances the fidelity of 3D scene reconstructions.

\subsection{Dynamic Region Density Control}

\begin{figure}[htb]
\begin{minipage}[b]{1.0\linewidth}
  \centering
  \includegraphics[scale=0.5]{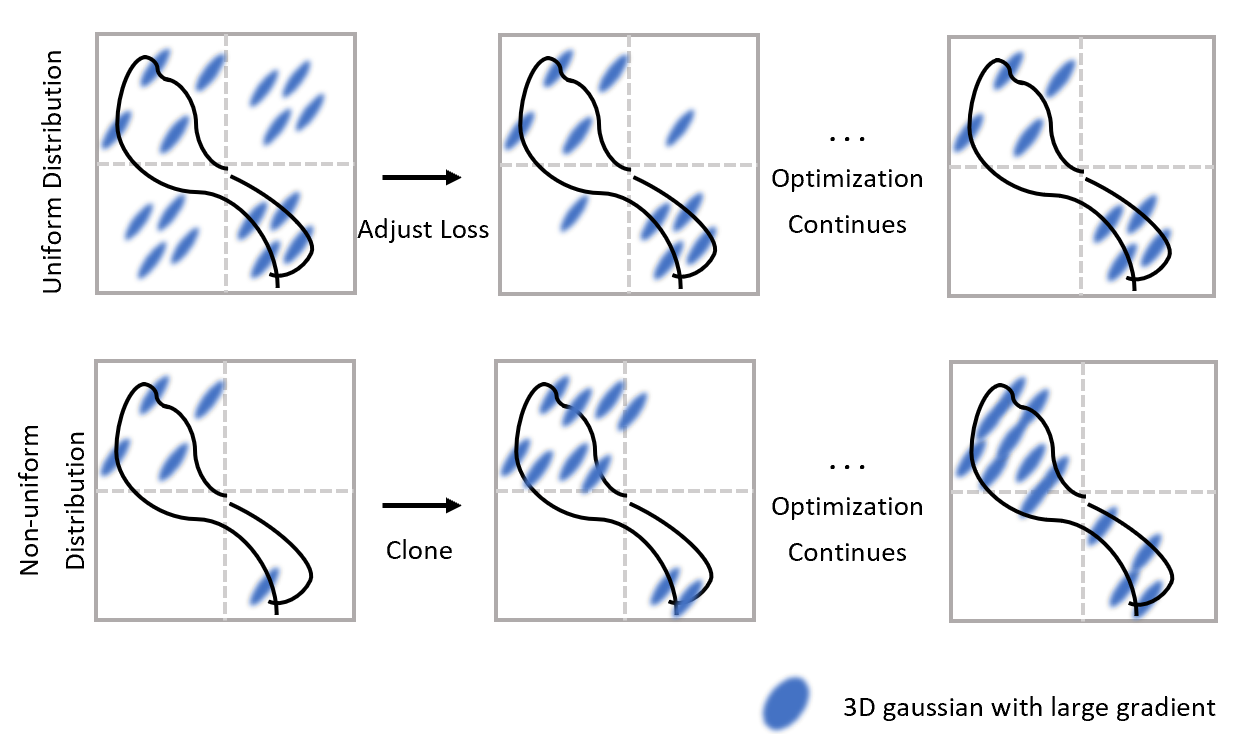}
  \caption{Loss adjustments in uniform regions  enhance Gaussian distributions, improving flexibility and computational efficiency. By restricting cloning to the Non-Uniform areas, the system increases visual details in high density area only}
  \label{fig:p3}
\end{minipage}
\end{figure}

The original Adaptive Density Control (ADC) system adjusts Gaussian density through a two-step strategy. First, transparency-based pruning identifies Gaussians with low transparency and close proximity to the camera as redundant and removes them. This reduces foreground clutter, optimizes memory usage, and enhances rendering clarity without compromising critical visual details. Mathematically, a Gaussian at position \(\mathbf{x}\) near the camera position \(\mathbf{c}\) is pruned if 
\begin{equation}
    \alpha(\mathbf{x}) < \delta_d \quad \text{and} \quad d(\mathbf{x}, \mathbf{c}) < \delta_d,
\end{equation}
where \(\alpha(\mathbf{x})\) represents transparency, \(d(\mathbf{x}, \mathbf{c})\) is the Euclidean distance to the camera, and \(\delta_d\) is the pruning threshold. Second, gradient-based cloning addresses regions requiring enhanced resolution by cloning Gaussians with high gradients \(\nabla I(\mathbf{x})\) and large sizes (defined by covariance matrix \(\Sigma\)). Cloning increases density in these areas, preserving fidelity without overloading simple regions. Cloning occurs if
\begin{equation}
    \nabla I(\mathbf{x}) > \delta_g \quad \text{and} \quad \text{size}(\Sigma) > \delta_s,
\end{equation}
where \(\delta_g\) and \(\delta_s\) are thresholds for gradients and Gaussian sizes, respectively. While effective, this approach applies uniform adjustments across regions, neglecting spatial variations in detail requirements, leading to uneven Gaussian distributions, visual artifacts, or reduced computational efficiency.

The Dynamic ADC system builds upon these principles by incorporating \textit{Top 20 Loss} to improve region-based Gaussian adjustments and distribution control as shown in Figure~\ref{fig:p3}. The scene is divided into fixed regions \(\mathcal{R}_k\) (\(k = 1, 2, \dots, N\)), enabling localized evaluation and adjustment of Gaussian distributions. Using \textit{Top 20 Loss}, the system prioritizes high-density regions, defined by the density ratio of the Top 20 densest to Bottom 20 sparsest sectors, ensuring critical areas receive more precise adjustments. Each region contains a subset of high-gradient Gaussians:
\begin{equation}
    \mathcal{G}_k = \{ \mathbf{x} \in \mathcal{R}_k \mid \nabla I(\mathbf{x}) > \delta_g \}.
\end{equation}
Regions with significant density variance \(\sigma_k^2\) exceeding the non-uniformity threshold \(\delta_u\) are classified as non-uniform. These regions are adjusted by cloning Gaussians, while uniform regions (\(\sigma_k^2 \leq \delta_u\)) are optimized through \textit{Top 20 Loss}, promoting even distributions and reducing redundancy.

For non-uniform regions (\(\sigma_k^2 > \delta_u\)), Gaussians are cloned to ensure adequate coverage. New Gaussians are positioned with small perturbations to maintain appropriate spatial distribution, targeting areas identified by the \textit{Top 20 Loss} mechanism. For uniform regions (\(\sigma_k^2 \leq \delta_u\)), adjustments are applied to the loss function \(\mathcal{L}_k\), specifically the dispersion term \(\mathcal{L}_{\text{top20}}\), to encourage more even Gaussian distributions. The dispersion term is defined as:
\begin{equation}
    \mathcal{L}_{\text{top20}} = \frac{1}{|\mathcal{G}_{\text{top20}}|} \sum_{\mathbf{x}_i, \mathbf{x}_j \in \mathcal{G}_{\text{top20}}} \|\mathbf{x}_i - \mathbf{x}_j\|^2,
\end{equation}
where \(\mathcal{G}_{\text{top20}}\) represents the set of Gaussians within the Top 20 densest regions. This term minimizes clustering, ensuring better spatial coverage.

The Dynamic ADC system significantly enhances both efficiency and fidelity. High-gradient regions prioritized by \textit{Top 20 Loss} receive finer Gaussian distributions, improving fidelity in detailed areas while avoiding redundant placement in simpler regions. By leveraging density ratios as a guiding metric, the system dynamically adapts to scene complexities, balancing high-frequency detail in critical areas with efficient resource allocation across the scene.

This updated ADC framework transitions from a uniform gradient-transparency model to a region-sensitive control mechanism. 

The Dynamic ADC system incorporates a composite loss function to achieve dynamic Gaussian distribution control, defined as:
\begin{equation}
    \mathcal{L}_{\text{ADC}} = \mathcal{L}_{\text{recon}}(\mathcal{R}_k) + \lambda_{\text{top20}} \mathcal{L}_{\text{top20}},
\end{equation}
where \(\mathcal{L}_{\text{recon}}(\mathcal{R}_k)\) represents the reconstruction loss for region \(\mathcal{R}_k\), and \(\lambda_{\text{top20}}\) adjusts the weight of \textit{Top 20 Loss} based on the region’s detail requirements.

 By incorporating region segmentation, gradient-sensitive adjustments, and \textit{Top 20 Loss}, the framework achieves a refined balance between image quality and computational performance, ensuring robust adaptability for high-fidelity rendering in complex 3D scenes.

\section{Implementation, Results, and Evaluation}

\begin{table*}[t]
    \centering
    \caption{Quantitative evaluation of our method compared to previous work, computed across Indoor, Outdoor, All Mip360 scenes, Tanks \& Temples (T\&T), and Deep Blending (DB) scenes.}
    \scalebox{0.7}{
    \begin{tabular}{|l|c|c|c||c|c|c||c|c|c||c|c|c||c|c|c|}
        \hline
        & \multicolumn{3}{c||}{\textbf{Indoor Scenes}} & \multicolumn{3}{c||}{\textbf{Outdoor Scenes}} & \multicolumn{3}{c||}{\textbf{Average Across All Scenes}} & \multicolumn{3}{c||}{\textbf{Tanks \& Temples (T\&T)}} & \multicolumn{3}{c|}{\textbf{Deep Blending (DB)}} \\
        \hline
        \textbf{Method} & \textbf{PSNR} & \textbf{LPIPS} & \textbf{SSIM} & \textbf{PSNR} & \textbf{LPIPS} & \textbf{SSIM} & \textbf{PSNR} & \textbf{LPIPS} & \textbf{SSIM} & \textbf{PSNR} & \textbf{LPIPS} & \textbf{SSIM} & \textbf{PSNR} & \textbf{LPIPS} & \textbf{SSIM} \\
        \hline
        Plenoxels \cite{FridovichKeil2022} & 24.83 & 0.426 & 0.766 & 22.02 & 0.465 & 0.542 & 23.62 & 0.443 & 0.670 & 21.08 & 0.379 & 0.719 & 23.06 & 0.510 & 0.795 \\
        INGP-Base \cite{Muller2022} & 28.65 & 0.281 & 0.840 & 23.47 & 0.416 & 0.571 & 26.43 & 0.339 & 0.725 & 21.72 & 0.330 & 0.723 & 23.62 & 0.423 & 0.797 \\
        INGP-Big \cite{Muller2022} & 29.14 & 0.242 & 0.863 & 23.57 & 0.375 & 0.602 & 26.75 & 0.299 & 0.751 & 21.92 & 0.305 & 0.745 & 24.96 & 0.390 & 0.817 \\
        Mip-NeRF360 \cite{Barron2022} & 31.58 & 0.182 & 0.914 & 25.79 & 0.247 & 0.746 & 29.09 & 0.210 & 0.842 & 22.22 & 0.257 & 0.759 & 29.40 & 0.245 & 0.901 \\
        3DGS-7K & 28.95 & 0.222 & 0.901 & 23.70 & 0.321 & 0.668 & 26.32 & 0.272 & 0.785 & 21.20 & 0.280 & 0.767 & 27.78 & 0.317 & 0.875 \\
        3DGS-30K & 31.05 & 0.186 & 0.925 & 24.69 & 0.239 & 0.729 & 27.87 & 0.213 & 0.827 & 23.14 & 0.183 & 0.841 & 29.41 & 0.243 & 0.903 \\
        ours-7K & 30.06 & 0.196 & 0.911 & 24.07 & 0.269 & 0.697 & 26.71 & 0.238 & 0.794 & 22.7 & 0.220 & 0.804 & 28.65 & 0.277 & 0.885 \\
        ours-30K & 31.72 & 0.179 & 0.925 & 25.16 & 0.221 & 0.739 & 28.36 & 0.191 & 0.834 & 24.39 & 0.151 & 0.854 & 30.28 & 0.230 & 0.903 \\
        \hline
    \end{tabular}
    }
    \label{tab:performance_comparison_combined}
\end{table*}

\begin{figure}[htb]
\begin{minipage}[b]{1.0\linewidth}
  \centering
  \includegraphics[scale=0.3]{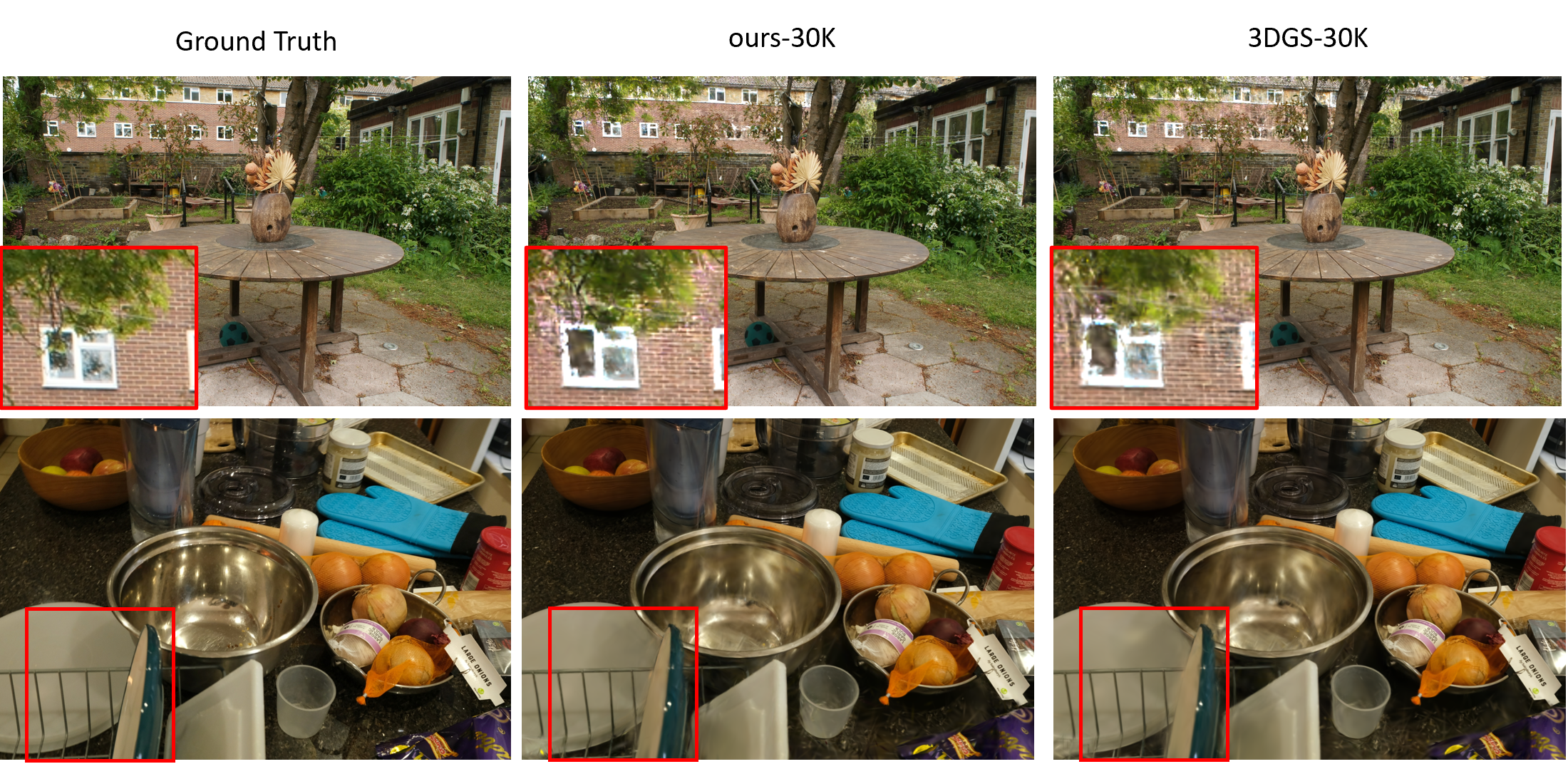}
  \caption{Qualitative comparison of reconstructed scenes for Ground Truth, ours-30K, and 3DGS-30K. The insets highlight fine details in both indoor and outdoor scenes, showing the superior reconstruction fidelity of our approach in preserving structural details}
  \label{fig:img}
\end{minipage}
\end{figure}
\begin{table}[t]
    \centering
    \setlength{\tabcolsep}{1pt}
    \caption{Ablation Results for Different Scenarios Across Datasets}
    \scalebox{0.74}{
    \begin{tabular}{|l|ccc|ccc|ccc|}
        \hline
        \textbf{Dataset} & \multicolumn{3}{c|}{\textbf{Mip360}} & \multicolumn{3}{c|}{\textbf{Deep Blending}} & \multicolumn{3}{c|}{\textbf{Tanks \& Temples}} \\
        \hline
        \textbf{Method} & \textbf{PSNR} & \textbf{SSIM} & \textbf{LPIPS} & \textbf{PSNR} & \textbf{SSIM} & \textbf{LPIPS} & \textbf{PSNR} & \textbf{SSIM} & \textbf{LPIPS} \\
        \hline
        NoInit  & 27.93 & 0.829 & 0.197 & 29.43 & 0.902 & 0.237 & 23.62 & 0.841 & 0.175 \\
        NoGloss & 28.29 & 0.832 & 0.196 & 30.13 & 0.894 & 0.241 & 24.37 & 0.848 & 0.168\\
        NoDynADC & 28.18 & 0.832 & 0.193 & 29.89 & 0.896 & 0.243 & 23.74 & 0.852 & 0.171 \\
        \hline
    \end{tabular}
    }
    \label{tab:ablation_results}
\end{table}

\subsection{Implementation}
All models are optimized on a single A800 GPU with 80 GB of memory. Training is divided into three stages across 30,000 iterations for efficient and accurate Gaussian splatting. Training spans 30,000 iterations divided into three phases. During the initial phase from 0 to 5,000 iterations, no regularization is applied, allowing Gaussians to adapt freely while L1 and SSIM loss ensure alignment. In the regularization phase from 5,000 to 30,000 iterations, a distance and orientation loss is applied every 100 iterations to refine alignment, with depth regularization included when depth maps are available. The density control phase from 10,000 to 30,000 iterations clones additional Gaussians in high-density areas to maintain fidelity. Link to code is available(https://github.com/ssssour/gd-3dgs).

\subsection{Results and Evaluation}

We evaluated several variations of our method on the same datasets as 3DGS, including Mip-NeRF360 \cite{Barron2022}, Tanks \& Temples \cite{Knapitsch2017}, Deep Blending \cite{Hedman2018}, and the synthetic Blender dataset \cite{Mildenhall2020}. The tests were carried out with consistent hyperparameters. Visual quality was assessed using standard metrics (PSNR, SSIM, and LPIPS), and our method demonstrated an improved degree of structural similarity while reducing pixel-wise discrepancies, achieving superior accuracy in capturing intricate details of the scene, especially in high-detail regions than 3DGS and other state-of-the-art methods. 

As shown in Table~\ref{tab:performance_comparison_combined}, our approach achieved high-quality results in as little as 7K iterations, with further improvements observed after 30K iterations. As shown in Figure~\ref{fig:img}, visual comparisons showed that our method reduced background artifacts and enhanced fine details, such as straight lines and distant window in outdoor scenes. Our method demonstrates high pixel-wise accuracy and structural integrity with minimal perceptual distortion. Compared to 3DGS-30K, it achieves improved perceptual quality, as seen in the higher SSIM and lower LPIPS values across diverse datasets, including Tanks \& Temples and Deep Blending. As shown in the quantitative tables, it excels in high-detail indoor scenarios with superior PSNR and SSIM scores and exhibits strong adaptability across diverse scenes, effectively handling varying complexities. Furthermore, it illustrates that our approach rivals or outperforms leading methods, particularly in perceptual quality, structural preservation, and robustness in dynamic scenarios.

\subsection{Ablations}
Ablation studies were conducted to evaluate the contributions of our innovations. As shown in Table ~\ref{tab:ablation_results}, the absence of geometric initialization led to degraded quality, particularly in background regions, highlighting the importance of structured initialization for stability as it impacts all three metrics. This underscores its critical role in maintaining scene quality across datasets. The removal of region-aware adjustments in adaptive density control resulted in uneven Gaussian distributions, mainly decreasing PSNR due to suboptimal density allocation but slightly improving SSIM and LPIPS, enhancing perceptual quality in simpler areas. Disabling surface-aligned optimization primarily worsened LPIPS while leaving PSNR and SSIM largely unaffected, demonstrating its significance for perceptual rendering quality, though its absence did not affect visual results significantly. These findings collectively highlight the importance of each component in optimizing high-complexity areas, maintaining uniformity in simpler regions, and ensuring overall scene stability and quality.

\section{Conclusions}
Our enhanced 3D Gaussian Splatting method addresses limitations in initialization, optimization, and density control, maintaining real-time performance. By using SfM data and an MLP for initialization, we achieve improved convergence and accuracy over the original 3DGS. Our approach refines Gaussian placement for higher fidelity reconstructions and introduces adaptive density control to optimize resource allocation. This method challenges continuous representations, showing explicit methods can achieve high-quality rendering with reduced training times and greater efficiency. Despite advancements, our research identifies areas for improvement, such as reducing GPU memory usage and potential for mesh reconstruction applications.

\bibliographystyle{IEEEbib}
\bibliography{icme2022template}

\end{document}